# FeaStNet: Feature-Steered Graph Convolutions for 3D Shape Analysis

Nitika Verma    Edmond Boyer    Jakob Verbeek

Univ. Grenoble Alpes, Inria, CNRS, Grenoble INP, LJK, 38000 Grenoble, France

`firstname.lastname@inria.fr`

## Abstract

*Convolutional neural networks (CNNs) have massively impacted visual recognition in 2D images, and are now ubiquitous in state-of-the-art approaches. CNNs do not easily extend, however, to data that are not represented by regular grids, such as 3D shape meshes or other graph-structured data, to which traditional local convolution operators do not directly apply. To address this problem, we propose a novel graph-convolution operator to establish correspondences between filter weights and graph neighborhoods with arbitrary connectivity. The key novelty of our approach is that these correspondences are dynamically computed from features learned by the network, rather than relying on predefined static coordinates over the graph as in previous work. We obtain excellent experimental results that significantly improve over previous state-of-the-art shape correspondence results. This shows that our approach can learn effective shape representations from raw input coordinates, without relying on shape descriptors.*

## 1. Introduction

In recent years, deep learning has dramatically improved the state of the art in several research domains including computer vision, speech recognition, and natural language processing [13]. In particular, convolutional neural networks (CNNs) have now become ubiquitous in computational solutions to visual recognition problems such as image classification [8], semantic segmentation [34], object detection [21], and image captioning [32]. CNNs also extend beyond 2D visual information, and easily generalize to other data that come in the form of regular rectangular grids. This has been demonstrated with for instance 1D convolution for audio signal [18] and 3D convolution over space and time for video signal [28].

Of particular interest beyond 2D image understanding are 3D shape models for which two main categories of representations can be considered. Extrinsic or Eulerian rep-resentations are based on parametrizations external to the shape, the most common being voxel grids. Such representations enable standard CNNs to be applied over 3D grids, but lack invariance to even the most basic transformations of the shape. A simple rigid transformation of the shape can lead to significant changes in the 3D grid representation. Moreover, discretizing space instead of shapes tends to be inefficient, in particular with moving and deforming objects for which a significant part of the space grid can be empty, resulting in representations with poor shape resolutions [7], or requiring special data structures to handle sparse inputs and/or outputs [22, 26]. On the other hand, intrinsic or Lagrangian representations, for example 3D meshes or volumetric quantizations, are robust to many shape transformations and describe 3D entities more efficiently with discretizations that are attached to shapes and not to the surrounding space. CNNs are, however, not readily extended to such representations with irregular structures represented as graphs where nodes can have a varying number of neighbors. The challenge is to define convolution-like operators over irregular local supports which can be used as layers in deep networks for prediction tasks such as shape correspondence over 3D meshes, see Figure 1.

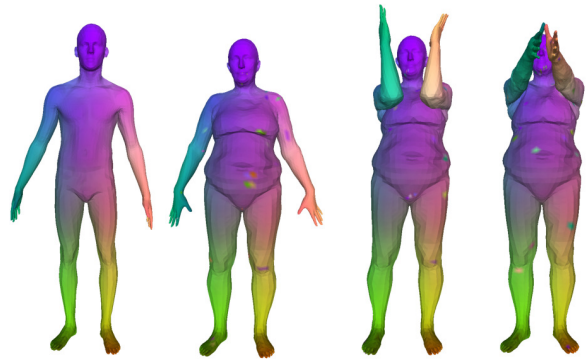

Figure 1. Three examples of texture transfer from a reference shape in neutral pose (left) using shape correspondences predicted by FeaStNet (multi-scale architecture, without refinement).

A number of architectures that generalize beyond data organized in regular grids have been recently proposed [2, 4, 5, 9, 11, 12, 14, 15, 20, 25]. Some of these techniques generalize beyond 3D shape data to other domains where data can be organized into graph structures, including for instance social networks or molecular graphs [4, 11]. The existing approaches come however with several limitations. Spectral filtering approaches [4, 5, 9, 11] rely on the eigen-decomposition of the graph Laplacian. Unfortunately this decomposition is often unstable, making the generalization across different shapes difficult [15]. Local filtering approaches [2, 14, 15] on the other hand, rely on possibly suboptimal hard-coded local pseudo-coordinates over the graph to define filters. Other approaches rely on point-cloud representations [12, 20] which cannot leverage surface information encoded in meshes, or need ad-hoc transformations of mesh data to map it to the unit sphere [25].

In this paper we present FeaStNet, a deep neural network based on a novel graph convolution operator which, unlike previous work, does not rely on static pre-defined local pseudo-coordinate systems over the graph, but instead uses the learned features of the preceding network layer to dynamically determine the association between filter weights and the nodes in a local graph neighborhood. Excellent experimental results on the FAUST 3D shape correspondence benchmark validate our approach, and significantly improve over recent state-of-the-art approaches. Figure 1 shows several examples of texture transfer using correspondences predicted with our model. Importantly, these results were obtained with the raw 3D shape coordinates as input instead of 3D shape descriptors as traditionally used for shape correspondence estimation. They demonstrate that FeaStNet learns better local shape properties than existing engineered 3D descriptors. Additional results on shape part labeling over point clouds are comparable to the state of the art, and illustrate that our approach generalizes to 3D data without explicit surface information.

## 2. Related work

In this section we briefly review related work on graph-convolutional networks, other deep learning approaches to process 3D shapes, and CNNs with data-adaptive filters.

**Graph-convolutional networks.** Existing approaches to generalize convolutional networks to irregular graph-structured data can be divided into two broad categories: spectral filtering methods and local filtering methods. Spectral methods build on a mathematically elegant approach to develop convolution-like operators using the spectral eigen-decomposition of the graph Laplacian [4, 5, 9, 11]. Any function defined over the graph nodes, *e.g.* features, can be mapped, by projection on the eigenvectors of the Laplacian, to the spectral domain where filtering consists of scaling the signals in the eigenbasis. While successful with noise-free data such as synthetic 3D shape models, spectral techniques are less suitable for acquired real shapes since global decompositions are unstable across different graphs, encoding for instance different shape meshes in various poses.

In an effort to better generalize across graphs, a number of techniques follow a strategy based on local graph filtering [2, 14, 15, 16, 24]. These methods differ in how they establish a correspondence between filter weights and nodes in local graph neighborhoods. Niepert *et al.* [16] rely on a heuristic ordering of the nodes, and then apply 1D CNNs. The geodesic CNN model of Masci *et al.* [14] extracts local patches on meshes which are convolved with filters expressed in polar coordinates. The orientation ambiguity of filters is dealt with by means of angular max-pooling, *i.e.* filters are applied in all possible orientations, and the maximum responses are retained. Boscani *et al.* [2] proposed the anisotropic CNN model which further extends the geodesic CNN model by using an anisotropic patch-extraction method, exploiting the maximum curvature directions to orient patches. Monti *et al.* [15] also parameterize local patches of the graph using fixed local polar pseudo-coordinates around each node. They learn filter shapes by estimating the means and variances of Gaussians that associate filter weights to the local pseudo-coordinates. Simonovsky & Komodakis [24] use edge labels, which play a similar role as the local pseudo coordinates, as input to a filter-generating subnetwork. Our work is related, though instead of relying on hand-designed local pseudo-coordinates, we learn the mapping between local graph patches and filter weights using the features in the previous network layer.

**Deep networks for 3D shape data.** Besides spectral and local filtering approaches on graphs, a number of other techniques have been developed to handle 3D shape data in deep neural networks. Sinha *et al.* [25] use a spherical parametrization, filling holes in the mesh when needed, to map shapes onto octahedra. These octahedra are cut and unfolded to square images, which can then be processed using regular CNNs. Wei *et al.* [30] render depth maps of shapes, and process them with conventional CNNs to learn features that can be matched to establish shape correspondence. Contrary to these approaches which transform 3D shape input data into 2D images that are fed to conventional CNNs, we propose instead a novel graph convolution that can directly process irregular graph-structured data.

Recently, two architectures have been proposed to process point cloud data. Klokov & Lempitsky [12] propose a deep network based on kd-trees over 3D point clouds, sharing parameters across the tree based on the depth and direction of splits. Qi *et al.* [19, 20] combine local per-point processing layers, with max-pooling layers to process 3D point clouds. By construction, these approaches ignore the surface information available in mesh data, and require sufficiently dense sampling to avoid artifacts due to spatial proximity of points that are geodesically remote.

**Data-adaptive convolutional networks.** The convolutional layers in a conventional CNN multiply together activations of the preceding feature map and learned filter weights, and sum the results to obtain the output as a linear function of the input, after which a point-wise non-linearity is applied. In spatial transformer networks [10] and dynamic filter networks [3], a subnetwork, which takes the preceding feature map as input, replaces a standard convolutional layer with a data-adaptive transformation. In the former, a localization subnetwork computes the parameters of a spatial transformation, *e.g.* cropping or re-sizing, which is used to spatially re-sample the preceding feature map before convolution. In the latter, a subnetwork is used to generate the convolutional filters that will be applied to the preceding feature maps. Our approach is related in the sense that we use a subnetwork to associate elements of a local "patch" of the graph to the filter weights.

## 3. Graph convolutions using dynamic filters

In this section we briefly revisit conventional CNNs, and then present our graph-convolutional network. We also compare the number of parameters and computational cost of our network with those of conventional CNNs.

### 3.1. Reformulating convolutional CNN layers

A convolutional CNN layer maps $D$ input feature maps to $E$ output feature maps. The parameters are commonly represented as a set of $D \times E$ filters $\mathbf{F}_{d,e}$, each of size $h \times w$ pixels, with $d \in \{1, \ldots, D\}$ and $e \in \{1, \ldots, E\}$. The computations in the convolutional layer to compute one of the $E$ output channels can be described as convolving each of the $D$ input channels with the corresponding filters, summing the $D$ convolution results and adding a constant bias to compute the output feature map.

An equivalent but less common representation, is useful to develop extensions for irregular graph-structured data. We rearrange the convolutional filter weights in a set of $M = h \times w$ weight matrices $\mathbf{W}_m \in \mathbb{R}^{E \times D}$. Each of these weight matrices is used to project input features $\mathbf{x} \in \mathbb{R}^D$ to output features $\mathbf{y} \in \mathbb{R}^E$. The result of the convolution at a pixel is obtained by summing for each of the $M$ neighbors the projection of its feature vector with the $\mathbf{W}_m$ corresponding to its relative position, considering pixel $i$ a neighbor of itself. See Figure 2 for an illustration. The activation $\mathbf{y}_i \in \mathbb{R}^E$ of pixel $i$ in the output feature map is written as

$$\mathbf{y}_i = \mathbf{b} + \sum_{m=1}^{M} \mathbf{W}_m \mathbf{x}_{n(m,i)}, \quad (1)$$

where $\mathbf{b} \in \mathbb{R}^E$ denotes a vector of bias terms, and $n(m, i)$ gives the index of the neighbor in the $m$-th relative position

w.r.t. pixel $i$. For example, the indices $n(1, i), \ldots, n(9, i)$ may refer to the pixels in a $3 \times 3$ patch centered at pixel $i$.

### 3.2. Generalization to non-regular input domains

In the case of CNNs for regular inputs, *e.g.* pixel grids, there is a clear one-to-one mapping between the weight matrices $\mathbf{W}_m \in \mathbb{R}^{E \times D}$ and the neighbors at relative positions $m \in \{1, \ldots, M\}$ w.r.t. the central pixel of the convolution. The main challenge in the case of irregular data graphs is to define this correspondence between neighbors and weight matrices. We propose to establish this correspondence in a data-driven manner, using a function over features computed in the preceding layer of the network, and learning the parameters of this function as a part of the network.

Instead of assigning each neighbor $j$ of a node $i$ to a single weight matrix, we use a soft-assignment $q_m(\mathbf{x}_i, \mathbf{x}_j)$ of the $j$-th neighbor across all the $M$ weight matrices. Given these soft-assignments, we generalize Eq. (1) and define the function that maps the features from one layer to the next as

$$\mathbf{y}_i = \mathbf{b} + \sum_{m=1}^{M} \frac{1}{|\mathcal{N}_i|} \sum_{j \in \mathcal{N}_i} q_m(\mathbf{x}_i, \mathbf{x}_j) \mathbf{W}_m \mathbf{x}_j, \quad (2)$$

where $q_m(\mathbf{x}_i, \mathbf{x}_j)$ is the assignment of $\mathbf{x}_j$ to the $m$-th weight matrix, and $\mathcal{N}_i$ is the set of neighbors of $i$ (including $i$), and $|\mathcal{N}_i|$ its cardinal.

We define the weights using a soft-max over a linear transformation of the local feature vectors as

$$q_m(\mathbf{x}_i, \mathbf{x}_j) \propto \exp\left(\mathbf{u}_m^\top \mathbf{x}_i + \mathbf{v}_m^\top \mathbf{x}_j + c_m\right), \quad (3)$$

with $\sum_{m=1}^{M} q_m(\mathbf{x}_i, \mathbf{x}_j) = 1$, and $\mathbf{u}_m$, $\mathbf{v}_m$ and $c_m$ are the parameters of the linear transformation. The weights involved in the update of node $i$ sum to 1 regardless of the number of neighbors of a node, since $\sum_{j \in \mathcal{N}_i} \frac{1}{|\mathcal{N}_i|} \sum_{m=1}^{M} q_m(\mathbf{x}_i, \mathbf{x}_j) = \sum_{j \in \mathcal{N}_i} \frac{1}{|\mathcal{N}_i|} = 1$. Therefore, our formulation is robust to variations in the degree of the nodes. Instead of using a single linear transformation of the features in Eq. (3), more general transformations may be used, such as a multi-layer sub-network. Conventional CNNs over grid-graphs are recovered if $\forall_i |\mathcal{N}_i| = M$, and the assignments are binary, *i.e.* $q_m(\mathbf{x}_i, \mathbf{x}_j) \in \{0, 1\}$, based on the relative position of neighbors w.r.t. node $i$. In Figure 2 we illustrate the computations in a standard grid CNN and in our graph convolutional network.

In our experiments, $\mathcal{N}_i$ contains vertex $i$ and all vertices connected to $i$ by an edge, *i.e.* the first *ring* around vertex $i$. Our approach, however, enables using larger neighborhoods, *e.g.* up to ring $k \geq 2$ or all vertices up to a certain geodesic distance. This is analogous to filters with larger spatial support in conventional CNNs. Importantly, and in contrast to standard CNNs, the above formulation decouples the neighborhood size $|\mathcal{N}_i|$ from the number of weight matrices $M$, and thus the number of parameters. As a consequence, filters with larger supports do not necessarily increase the number of parameters. Rather than relying on dilation [34] or weight-sharing for large filters, our approach learns the mapping between weights and neighbors.

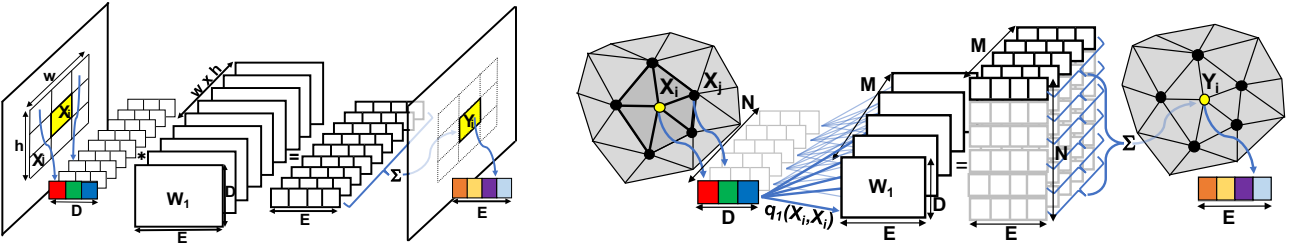

Figure 2. Left: Illustration of a standard CNN, representing the parameters as a set of $M = w \times h$ weight matrices, each of size $D \times E$. Each weight matrix is associated with a single relative position in the input patch. Right: Our graph convolutional network, where each node in the input patch is associated in a soft manner to each of the $M$ weight matrices based on its features using the weight $q_m(\mathbf{x}_i, \mathbf{x}_j)$.

**Translation invariant assignments in feature space.** As a special case, we can set $\mathbf{u}_m = -\mathbf{v}_m$ in Eq. (3), which results in $q_m^{ij} \propto \exp\left(\mathbf{u}_m^\top(\mathbf{x}_j - \mathbf{x}_i) + c_m\right)$, and leads to translation invariance of the weights in the feature space. This is of particular interest in applications where the input features include spatial coordinates, in which case it is natural to impose translation invariance on the assignment function. Our experimental results confirm the positive effect of translation invariance when using raw spatial 3D coordinates as input features for shape meshes.

**Assignment by Mahalanobis distance in feature space.** Another interesting case occurs when considering a Mahalanobis distance to determine the assignments weights $q_m(\mathbf{x}_i, \mathbf{x}_j)$. The Mahalanobis distance, parameterized by a positive definite matrix $\Sigma$, between reference points $\mathbf{z}_m$ and a centered version of the neighbor features $\mathbf{x}_{ij} = \mathbf{x}_j - \mathbf{x}_i$, is given by

$$d_\Sigma(\mathbf{x}_{ij}, \mathbf{z}_m) = (\mathbf{x}_{ij} - \mathbf{z}_m)^\top \Sigma (\mathbf{x}_{ij} - \mathbf{z}_m) \quad (4)$$
$$= -2\mathbf{x}_{ij}^\top \Sigma \mathbf{z}_m + \mathbf{z}_m^\top \Sigma \mathbf{z}_m + \mathbf{x}_{ij}^\top \Sigma \mathbf{x}_{ij}. \quad (5)$$

The soft-assignments based on the Mahalanobis distances fit the form of Eq. (3) with $c_m = \mathbf{z}_m^\top \Sigma \mathbf{z}_m$, $\mathbf{u}_m = -2\Sigma \mathbf{z}_m$, and $\mathbf{v}_m = -\mathbf{u}_m$. These soft-assignments may be interpreted as the posterior assignments of the neighbor's centered feature vectors $\mathbf{x}_{ij}$ over the components of a Gaussian mixture model in feature space with means $\mathbf{z}_m$ and shared covariance matrix $\Sigma^{-1}$.

This mixture model interpretation of the soft-assignments highlights the connection between our approach and that of Monti *et al*. [15]. In the latter, a similar formulation is used in which centers $\mathbf{z}_m$ are learned along with covariance matrices $\Sigma_m$. This mixture is, however, defined over a-priori defined local pseudo-coordinates $\mathbf{x}_{ij}$ over the graph, *e.g.* local polar coordinates over a mesh, rather than learned features as in our formulation.

Using this formulation, we can also recover conventional CNNs over pixel grids as a special case by letting the pixel coordinates be part of the feature vectors $\mathbf{x}$, having the Mahalanobis distance depend only on these coordinates, and placing the centers $\mathbf{z}_m$ precisely on the relative positions of the neighboring pixels. Multiplying the Mahalanobis distances by a large constant will recover the hard-assignments used in the standard CNN model of Eq. (1).

### 3.3. Complexity analysis

The weight matrices $\mathbf{W}_m$ are common between a conventional CNN and our approach, and contain $MDE$ parameters. The only additional parameters in our approach w.r.t. a conventional CNN are the vectors $\mathbf{u}_m, \mathbf{v}_m$, which contain $2MD$ parameters. Thus the total number of parameters increases only by a factor $(E+2)/E = 1 + 2/E$, ignoring bias terms which contribute very few parameters.

To efficiently evaluate the activations, we first multiply all feature vectors $\mathbf{x}_i$ with the weight matrices $\mathbf{W}_m$ and weight vectors $\mathbf{u}_m$, and $\mathbf{v}_m$. This takes $\mathcal{O}(NMDE)$ operations, where $N$ is the number of nodes in the graph. Let $K$ denote the average number of neighbors of each vertex, we can then compute the weights in Eq. (3) and the activations in Eq. (2) in $\mathcal{O}(NMKE)$ operations. The total computational cost is thus $\mathcal{O}(NME(K+D))$.

The cost of a convolutional layer in a conventional CNN is $\mathcal{O}(NMED)$, *c.f.* Eq. (1). The computational cost of our approach is thus comparable, provided the number of neighbors $K$ is comparable or smaller than the number of features $D$, as is typically the case in practice.

## 4. Experimental evaluation

We evaluate our approach on 3D shape correspondence between 3D meshes. In addition, we present results on part labeling of point cloud data, where we apply our model on ad-hoc neighborhood graphs.

### 4.1. 3D shape correspondence

**Experimental setup.** We follow the experimental setup in [2, 14, 15] based on the FAUST human shape dataset [1].

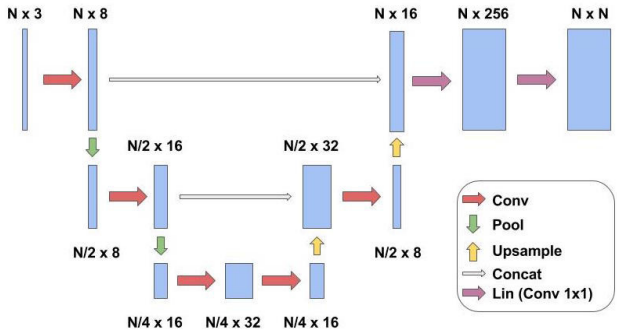

Figure 3. Our multi-scale graph convolution architecture.

This dataset consists of 100 watertight meshes with 6,890 vertices each, corresponding to 10 shapes in 10 different poses each. The shape correspondence problem, between a given reference shape and any other shape, is formulated as a vertex labeling problem where the label set consists of all the 6,890 vertices on the reference shape. The first 80 shape meshes are used as training data, and the last 20 meshes are used as test data, corresponding to the 10 poses of two shapes not seen during training. Exact ground-truth correspondence is known, and the first shape in the first pose is used as reference. The output of the last soft-max layer at each vertex gives a probability distribution over the corresponding point on the reference shape.

Unless specified otherwise, we follow the network architecture of [14], which is similar to the ones used in [2, 15]. It consists of the following sequence of linear layers ($1\times 1$ convolutions) and graph convolutions: Lin16+Conv32+Conv64+Conv128+Lin256+Lin6890; the numbers indicate the amount of output channels of each layer. In addition we developed a multi-scale architecture with pooling and unpooling layers inspired by U-Net [23], which increases the field of view without losing spatial resolution. Following Defferrard *et al*. [5], we use the Graclus algorithm [6] to define max-pooling over the graph. Given a graph with edge weights $w_{ij}$ and degrees $d_i = \sum_j w_{ij}$, this greedy clustering algorithm merges in each step the unmarked nodes that maximize the local normalized cut $w_{ij}(d_i^{-1} + d_j^{-1})$, and then marks these nodes as visited. For simplicity we set all initial edge weights equal to 1. The coarsened graph has approximately two times fewer nodes, and the weights in the coarsened graph are set to the sum of the corresponding weights before coarsening. This process is repeated to construct a binary tree over the nodes. This induces a complete ordering which can be used to apply standard 1-dimensional max-pooling layers, as well as fractionally strided convolution upsampling layers. Our multi-scale architecture is illustrated in Figure 3.

The models are trained using the standard cross-entropy classification loss. We use learning rate of $10^{-2}$, and a

| Translation inv. | yes | no |
|---|---|---|
| XYZ | 86% | 28% |
| SHOT | 63% | 58% |

Table 1. Shape correspondence accuracy using different input features, with and without translation invariance, $M = 9$ in all cases.

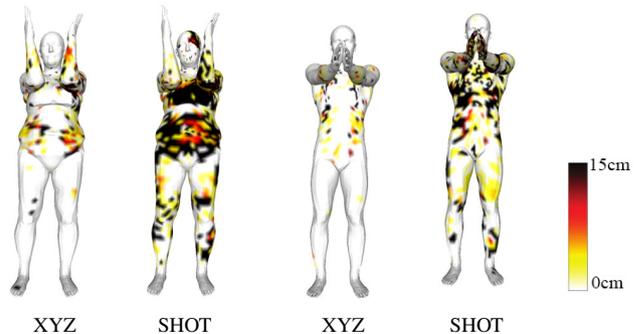

XYZ　　SHOT　　XYZ　　SHOT

Figure 4. Geodesic errors on two test shapes estimated using single-scale architecture with XYZ and SHOT feature inputs.

weight decay of $10^{-4}$. As input features over the mesh we either use the 544-dimensional SHOT descriptor [27] used in earlier work, or the raw 3D XYZ vertex coordinates. The accuracy is defined as the number of vertices for which the correspondence prediction is exact, but we also evaluate the number of correspondence predictions within a certain tolerance on the error in terms of geodesic distance.

**Results.** In Table 1 we evaluate our single-scale model using the XYZ coordinates and the SHOT descriptor as input, and with and without translation invariance in our model. For both descriptors translation invariance improves results. As expected translation invariance is more important in the case of raw XYZ inputs, since the coordinates have no built-in translation invariance while the SHOT descriptor is invariant to the absolute position of the local shape. With translation invariance, the XYZ inputs clearly outperform the SHOT descriptor, demonstrating that our model can learn shape features that outperform state-of-the-art hand-crafted shape descriptors. Unless specified otherwise, we use XYZ inputs and translation invariance in the remaining experiments. In Figure 4 we visualize geodesic correspondence errors for both descriptors, clearly showing superior results using the raw XYZ coordinate input.

We evaluate the impact of the number of weight matrices $M$ in Figure 5. We observe that the performance quickly improves from $M = 2$ to $M = 8$, after which the improvements are smaller. This shows that the internal features learned by our model are effective to steer the graph convolutions and to successfully assign different weight matrices across a graph neighborhood. We use $M = 32$ for the remaining experiments.

In Table 2, we present the accuracy obtained with FeaSt-

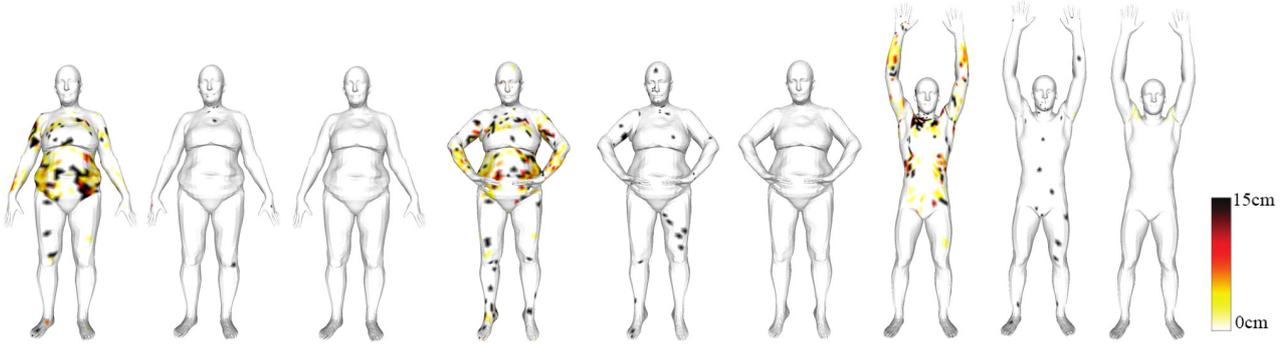

Figure 7. Visualization of correspondence errors in terms of the geodesic distance to the groundtruth correspondence on three test shapes, using (from left to right) the single-scale architecture (w/o refinement) and multi-scale architecture without and with refinement.

| Method | Input | Accuracy |
|---|---|---|
| Logistic Regr. | SHOT | 39.9% |
| PointNet [19] | SHOT | 49.7% |
| ACNN [2], w/o refinement | SHOT | 60.6% |
| ACNN [2], w/ refinement [17] | SHOT | 62.4% |
| GCNN [14], w/o refinement | SHOT | 65.4% |
| GCNN [14], w/ refinement | SHOT | 42.3% |
| MoNet [15], w/o refinement | SHOT | 73.8% |
| MoNet [15], w/ refinement [29] | SHOT | 88.2% |
| FeaStNet, w/o refinement | XYZ | 88.1% |
| FeaStNet, w/ refinement [29] | XYZ | 92.2% |
| FeaStNet, multi scale, w/o refinement | XYZ | 98.6% |
| FeaStNet, multi scale, w/ refinement [29] | XYZ | 98.7% |
| FeaStNet, multi scale, w/o refinement | SHOT | 90.9% |

Table 2. Correspondence accuracy on the FAUST dataset of our model and recent state-of-the-art approaches.

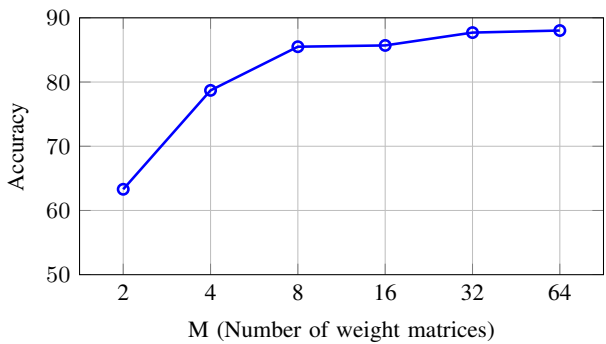

Figure 5. Accuracy as a function of the number of weight matrices for the FAUST dataset, using the single scale architecture.

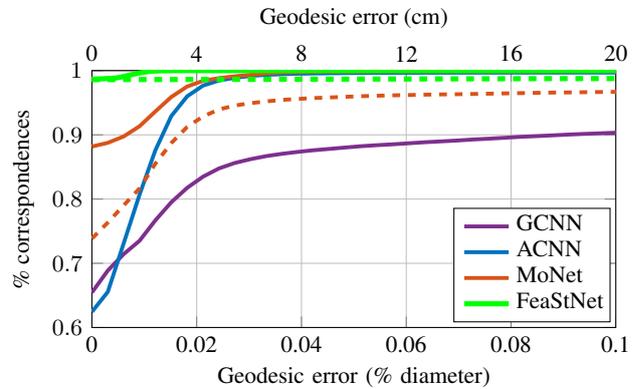

Figure 6. Fraction of geodesic shape correspondence errors within a certain distance. Dashed curves show results without refinement.

Net using the single-scale and multi-scale architecture, and compare to state-of-the-art methods. We also evaluate our best model (translation invariant multi-scale architecture) using SHOT descriptors, and obtain an accuracy significantly above previous state of the art. Accuracies for [2, 14, 15] are directly taken from the corresponding papers, and for PointNet we trained a model using the publicly available code. For sake of direct comparability, we evaluate the quality of the correspondences directly predicted by our model, and after post-processing them with the refinement algorithm of Vestner *et al*. [29] which was also used by Monti *et al*. [15]. Using our models we obtain excellent correspondence predictions. Our multi-scale architecture, which allows to use more contextual information across the mesh, predicts 98.6% of the correspondences without any error. In Figure 6 we plot the percentage of correspondences that are within a given geodesic distance from the ground truth on the reference shape. Figure 7 visualizes the geodesic correspondence errors using our single-scale and multi-scale architectures, and the effect of refinement. While refinement has only a marginal effect on the accuracy, it does in certain cases correct some of the rare relatively large errors. The correspondences predicted by our multi-scale network improve significantly over the previous state-of-the-art results of Monti *et al*. [15].

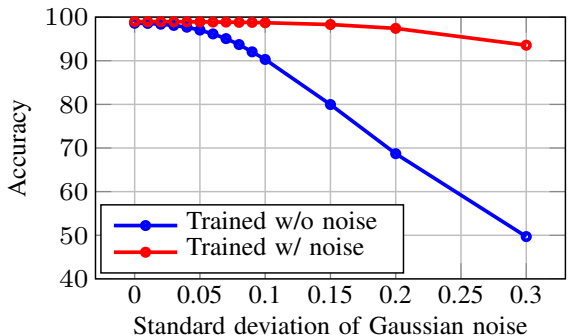
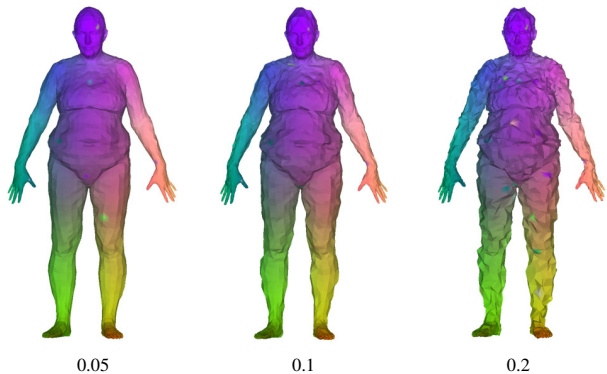

Figure 8. Left: Accuracy as a function of standard deviation of Gaussian noise added to FAUST test shapes. Right: Texture transfer on test shapes with various levels of additive Gaussian noise using our multi-scale FeaStNet architecture (trained with noisy data).

To evaluate the robustness of our models, we add Gaussian noise to each vertex of the shapes, where we use a locally adaptive standard deviation proportional to the local average inter-vertex distance. We visualize the results of multi-scale FeaStNet model on these new shapes in Figure 8. The blue curve demonstrates how the predictive performance deteriorates as the noise increases, when training on noise-free data. The red curve is obtained when also using noisy training data, using noise levels 0.01, 0.05, 0.1, 0.15 and 0.2. Adding noise to the training data can be seen as a form of data augmentation, and makes the model significantly more robust.

In Figure 9, we show activations of some randomly selected features learned across some layers our single-scale model. Across the layers the features become more pose invariant and more localized as required by the task.

## 4.2. Part labeling

**Experimental setup.** To validate our approach on graphs that are less clean than the ones in the FAUST dataset, we test it on the ShapeNet part benchmark [33]. The dataset consists of 16,881 shapes from 16 categories, labeled with 50 parts in total. Ground-truth labels are available on points sampled from the original shapes, but not on the original meshes themselves. Therefore, we apply our model on k-nearest neighbor graphs over the labeled 3D points. We follow the standard experimental protocol [12, 19, 31, 33], and report the mean intersection over union (mIoU) metric per category and across all shapes.

For each class a subset of the part labels is used, and the category labels are know for test shapes. Therefore, we train the model with a cross-entropy loss over the part labels corresponding to the category of each sample. We use $k = 16$ neighbors to construct the graph, and use $M = 16$ weight matrices. Our architecture consists of the following layers: Lin16-Conv32-Conv64-Conv128-Lin512-Lin2048-MaxPool. We concatenate the features from all the layers with the global max-pooled features, and feed them to two linear layers (Lin1024-Lin50) to get the final output.

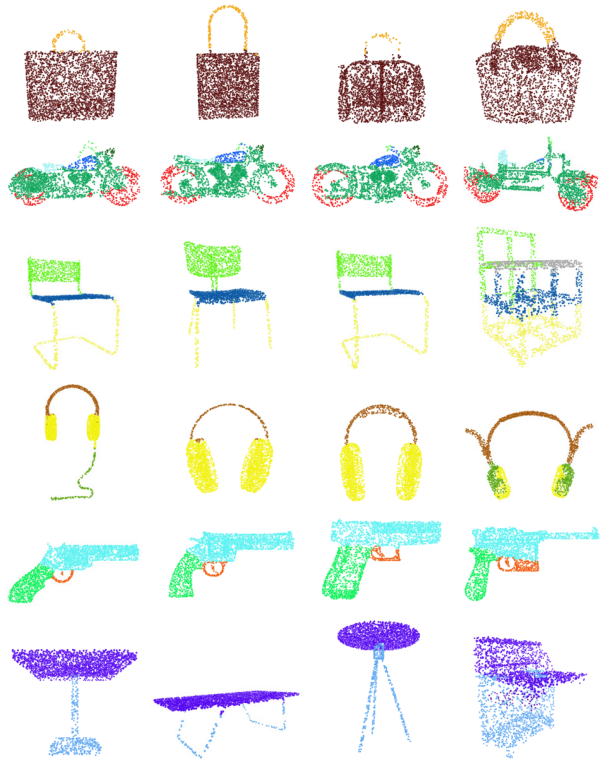

Figure 10. Part labeling results on ShapeNet. On each row we show three test shapes with accurate labeling, and one shape with the worst labeling in that category. Best viewed in color.

**Results.** The results in Table 3 show that we obtain results that are comparable to the state of the art. This demonstrates that our approach is not only effective on clean mesh graphs, but is also directly applicable to nearest neighbor graphs constructed from point clouds. We show results with number of nearest neighbors $k = 16$ as we did not observe much difference when varying $k$. In particular, we obtained mIoU of 79.9% ($k=4$), 80.8% ($k=8$), 81.5% ($k=16$), 80.9% ($k=32$). We provide part labeling results on several test shapes from seven categories in Figure 10. The failure cases mostly concern atypical shapes, *e.g.* for *table* and *chair*, and cases where the boundary between object labels is poorly estimated, *e.g.* for *bag*, *guitar* and *gun*.

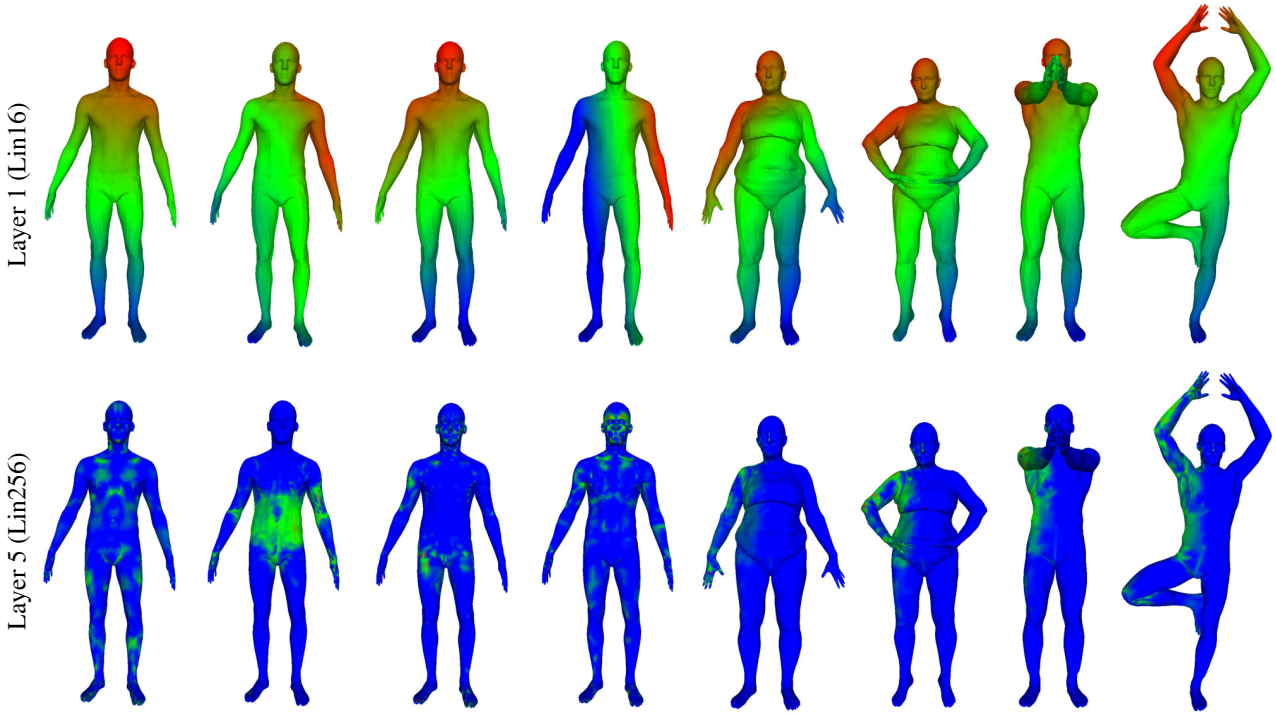

Figure 9. Visualization of activations of randomly selected features across first and last layers of our single-scale FeaStNet architecture using coordinates (xyz) as input. The first four columns show different features on a single shape, while the last four columns show another random feature across different shapes.

|  | overall | aero plane | bag | cap | car | chair | ear phone | guitar | knife | lamp | laptop | motor bike | mug | pistol | rocket | skate board | table |
|---|---|---|---|---|---|---|---|---|---|---|---|---|---|---|---|---|---|
| Number of shapes | 16,881 | 2690 | 76 | 55 | 898 | 3758 | 69 | 787 | 392 | 1547 | 451 | 202 | 184 | 283 | 66 | 152 | 5271 |
| Wu [31] | - | 63.2 | - | - | - | 73.5 | - | - | - | 74.4 | - | - | - | - | - | - | 74.8 |
| Yi [33] | 81.4 | 81.0 | 78.4 | 77.7 | 75.7 | 87.6 | 61.9 | 92.0 | 85.4 | 82.5 | 95.7 | 70.6 | 91.9 | 85.9 | 53.1 | 69.8 | 75.3 |
| PointNet [19] | 83.7 | 83.4 | 78.7 | 82.5 | 74.9 | 89.6 | 73.0 | 91.5 | 85.9 | 80.8 | 95.3 | 65.2 | 93.0 | 81.2 | 57.9 | 72.8 | 80.6 |
| Kd-network [12] | 82.3 | 80.1 | 74.6 | 74.3 | 70.3 | 88.6 | 73.5 | 90.2 | 87.2 | 81.0 | 94.9 | 57.4 | 86.7 | 78.1 | 51.8 | 69.9 | 80.3 |
| FeaStNet (this paper) | 81.5 | 79.3 | 74.2 | 69.9 | 71.7 | 87.5 | 64.2 | 90.0 | 80.1 | 78.7 | 94.7 | 62.4 | 91.8 | 78.3 | 48.1 | 71.6 | 79.6 |

Table 3. Part labeling accuracy in mIoU on the ShapeNet part dataset of our model and recent state-of-the-art approaches.

## 5. Conclusion

We presented FeaStNet, a novel graph-convolutional architecture which is based on local filtering and applies to generic graph structures, both regular and irregular. The main novelty is that our architecture determines local filters dynamically based on the features in the preceding layer of the network. The network thus learns features that are (i) effective to shape the local filters, and (ii) informative for the final prediction task. We obtain results that significantly improve over the state-of-the-art for 3D mesh correspondence on the FAUST dataset, and results comparable to the state of the art for part labeling on the ShapeNet dataset where we apply our model on k-nearest neighbor graphs over point clouds. In the future we plan to extend our architecture to model other properties of 3D shapes, such as appearance or motion patterns. The TensorFlow-based implementation to replicate our experiments can be found at: https://github.com/nitika-verma/FeaStNet

**Acknowledgment.** This work was in part supported by the French research agency contracts ANR16-CE23-0006 and ANR-11-LABX-0025-01.